\documentclass[conference]{IEEEtran}
\IEEEoverridecommandlockouts

\usepackage{hyperref}
\usepackage{xcolor}

\usepackage{times}
\usepackage{latexsym}
\usepackage{microtype}
\usepackage{booktabs}
\usepackage{textcomp}
\usepackage{amsmath}
\usepackage{graphicx}
\usepackage{tabulary}
\usepackage{pifont}
\usepackage{subcaption}
\usepackage{comment}
\usepackage{soul}

\usepackage{tabularx}
\usepackage{siunitx}
\usepackage{caption}

\usepackage[para]{footmisc}

\usepackage{adjustbox}
\usepackage{array}
\usepackage{xspace}

\newcommand{\ex}[1]{\textit{#1}\xspace}

\begin{document}

\title{Impact of detecting clinical trial elements in exploration of COVID-19 literature}
%
%
%
%


\author{\IEEEauthorblockN{Simon Šuster}
\IEEEauthorblockA{\textit{Fac. of Engineering and Information Tech.} \\
\textit{The University of Melbourne}\\
Melbourne, Australia \\
simon.suster@unimelb.edu.au}
\and
\IEEEauthorblockN{Karin Verspoor}
\IEEEauthorblockA{\textit{School of Computing Technologies} \\
\textit{RMIT University}\\
Melbourne, Australia \\
karin.verspoor@rmit.edu.au}
\and
\IEEEauthorblockN{Timothy Baldwin}
\IEEEauthorblockA{\textit{Fac. of Engineering and Information Tech.} \\\textit{The University of Melbourne}\\
Melbourne, Australia \\
tbaldwin@unimelb.edu.au}
\and
\IEEEauthorblockN{Jey Han Lau}
\IEEEauthorblockA{\textit{Fac. of Engineering and Information Tech.} \\\textit{The University of Melbourne}\\
Melbourne, Australia \\
jeyhan.lau@unimelb.edu.au}
\and
\IEEEauthorblockN{Antonio Jimeno Yepes}
\IEEEauthorblockA{\textit{Fac. of Engineering and Information Tech.} \\\textit{The University of Melbourne}\\
Melbourne, Australia \\
antonio.jimeno@gmail.com}
\and
\IEEEauthorblockN{David Martinez Iraola}
\IEEEauthorblockA{\textit{Independent researcher} \\
Melbourne, Australia \\
david.martinez.iraola@gmail.com}
\and
\IEEEauthorblockN{Yulia Otmakhova}
\IEEEauthorblockA{\textit{Fac. of Engineering and Information Tech.} \\
\textit{The University of Melbourne}\\
Melbourne, Australia \\
yotmakhova@student.unimelb.edu.au}
}
\maketitle

\begin{abstract}
 The COVID-19 pandemic has driven ever-greater demand for tools which enable efficient exploration of biomedical literature. 
 Although semi-structured information resulting from concept recognition and detection of the defining elements of clinical trials (e.g.\ PICO criteria) has been commonly used to support  literature search, the contributions of this abstraction remain poorly understood, especially in relation to text-based retrieval. In this study, we compare the results retrieved by a standard search engine with those filtered using clinically-relevant concepts and their relations. With analysis based on the annotations from the TREC-COVID shared task, we obtain quantitative as well as qualitative insights into characteristics of relational and concept-based literature exploration. Most importantly, we find that the relational concept selection filters the original retrieved collection in a way that decreases the proportion of unjudged documents and increases the precision, which means that the user is likely to be exposed to a larger number of relevant documents. 
 
\end{abstract}
\section{Introduction}
The outbreak of Coronavirus disease 2019 (COVID-19) has led to a vigorous response from the global medical and AI communities, with efforts in the fields of information retrieval and natural language processing revolving around dataset construction and the development of tools for managing the growing literature on the virus and related diseases \cite{Hutson2020}. The release of the COVID-19 Open Research Dataset (CORD-19) \cite{wang-etal-2020-cord} has stimulated the development of a large number of tools, reviewed in \cite{wang-lo-2020-bib}, which can be divided into more retrieval/QA-focused systems---returning a list of relevant documents for a user-specified query \cite{vespa,discovid,covidex,cord19,covidscholar}---and those that use domain knowledge to organise and present information found in the literature \cite{Hope2020SciSightCF,nye-etal-2020-trialstreamer,verspoor-etal-2021-brief,covidnavigator}.

Broad clinically-relevant PICO categories derived from the structure of clinical trials  \cite{richardson1995well} have been used to enable structured search as well as the visualisation of document content \cite{nye-etal-2020-trialstreamer,Hope2020SciSightCF,docsearch,covid-love,verspoor-etal-2021-brief}. These categories describe the \textbf{P}atient population
enrolled (e.g.\ \textit{diabetics}), the \textbf{I}nterventions studied (e.g.\ \textit{insulin}) and to
what they were \textbf{C}ompared (e.g.\ \textit{placebo}), and the \textbf{O}utcomes measured (e.g.\ \textit{blood glucose levels}).

\begin{figure*}[t]
    \centering
    \begin{subfigure}{1\textwidth}
    \centering
    \includegraphics[width=0.9\linewidth]{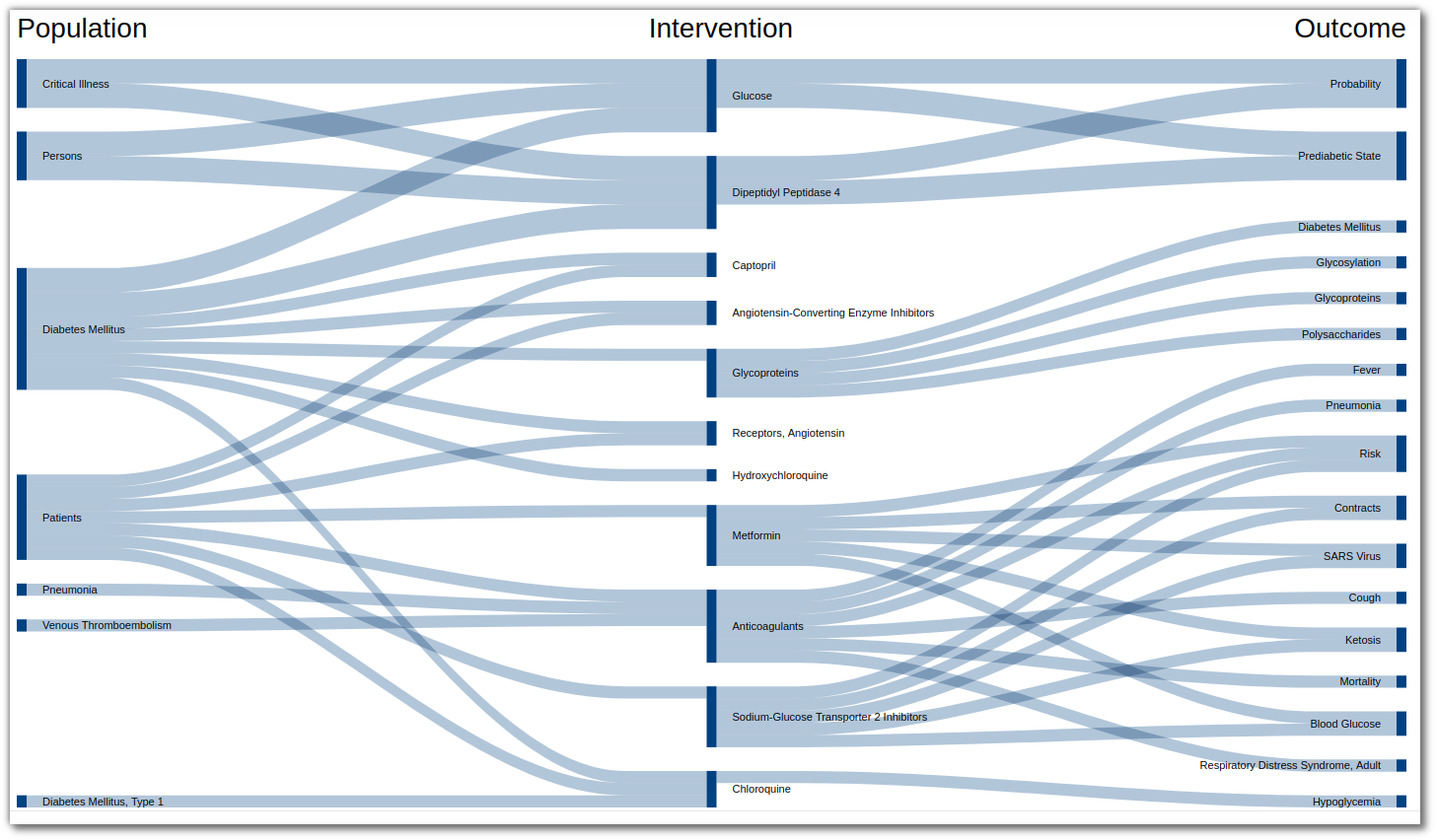}
    \end{subfigure}
    
    \begin{subfigure}{1\textwidth}
    \centering
    \includegraphics[width=0.9\linewidth]{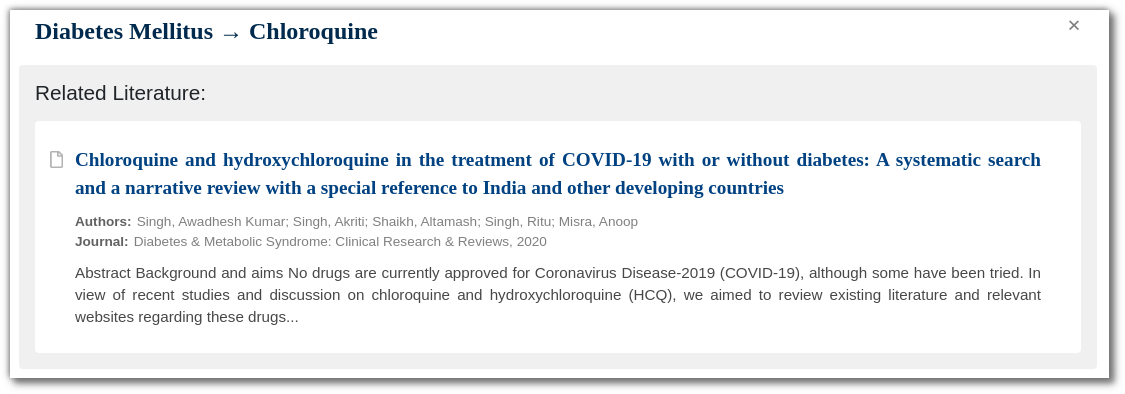}
    \end{subfigure}
    \caption{Sankey diagram of PICO concepts and relations for 35 articles retrieved for \ex{what kinds of complications related to COVID-19 are associated with diabetes} (TREC Round 5, query no.\ 24). Selecting a link (e.g.\ \ex{Diabetes Mellitus} $\rightarrow$ \ex{Chloroquine}) reveals papers that contain that relation (the displayed relation in the example only contains one document). Links are weighted by the number of documents containing the relevant relation.}
    \label{fig:sankey}
\end{figure*}

Here, we aim to examine the impact of PICO content structuring on literature exploration. 
Our analysis is based on the COVID-SEE system \cite{verspoor-etal-2021-brief,verspoor2020covid} which offers both simple document retrieval and diverse visualisations of the retrieved document collection. We specifically focus on the information presented in the relational concept selection of COVID-SEE, which adopts Sankey diagrams to visually organise the medical concepts found in the articles according to PICO categories. The aim of presenting the retrieved documents in this way is to highlight salient P--I and I--O concept relations\footnote{The comparator (`C') category is usually merged with interventions (`I') due to their high similarity for the purposes of automatic PICO labelling \cite{nye-etal-2018-corpus}. We also follow this practice here.}. 
Figure~\ref{fig:sankey} gives an example.

P--I and I--O relations often cannot be extracted from the retrieved documents, 
as not all articles are structured according to PICO and the identification of PICO elements is imperfect.
This means that the relational concept selection shown in the visual summary is \textit{implicitly filtering} a smaller subset of documents from the retrieval results. The question we seek to answer is how this affects the quality of the results. We approach this question by analysing the results through an existing test collection of relevance decisions, rather than carrying out a user study, which we leave for future work.
To obtain a ground truth for relevance of the retrieved documents, we use the judgements for COVID-19-related literature obtained in the TREC-COVID shared task \cite{roberts-etal-2020-trec}. 

Our findings reveal the following:

\begin{itemize}
    \item While the relational concept selection reduces the proportion of displayed documents to around 15\% of the base retrieval results,  the proportion of relevant documents actually \textit{increases}. 
    We find that this advantage comes from the fact that the relational concept selection is more likely to show the documents that received a judgment in the TREC-COVID test set (relevant or irrelevant). 
    \item Based on manually rating a sample of unjudged documents, we find that most are irrelevant, suggesting that PICO filtering boosts the overall quality of the search results.
    \item
    Furthermore, when calculating precision (the proportion of relevant documents to all retrieved documents), we observe that the results vary substantially across queries. This leads us to infer that queries which are more amenable to PICO representation---based on a manual annotation of PICO elements in the queries---also tend to correlate with higher precision in the relational concept selection.
    In a similar vein, we find that the strength of a P--I/I--O relation (i.e.\ the number of articles contained in that particular relation) is positively correlated with the proportion of relevant documents included. These results indicate that for medical queries with more easily identifiable PICO elements, 
    the use of these elements as a central organising principle in the COVID-SEE system leads to users being exposed to more relevant results.
\end{itemize}




\section{Empirical setup}
\subsection{Data}
In our experiments, we use the CORD-19 dataset (v.\ 16 July 2020, totalling around 192,000 documents). It is currently the most extensive publicly available corpus of coronavirus-related publications from different sources, including PubMed’s PMC open access corpus, research articles from a corpus maintained by the WHO, and bioRxiv and medRxiv pre-prints \cite{wang-etal-2020-cord}. 

\paragraph{\textbf{PICO-concept annotation}} 
To annotate PICO elements 
and biomedical concepts found within PICO phrases, we follow a two-step procedure described in detail in \cite{verspoor2020covid,verspoor-etal-2021-brief}. Briefly, in the first step, a BiLSTM-CRF model trained on the EBM-NLP dataset \cite{nye-etal-2018-corpus} is used to label textual spans with Population, Intervention, and Outcome categories. On the same dataset, the model's performance has been estimated at 0.78, 0.60 and 0.67 F1 for P, I and O categories, respectively.
The next step consists of applying  MetaMap \cite{metamap} to identify the terms that correspond to Medical Subject Headings (MeSH) \cite{mesh}. MeSH terms provide an established vocabulary for 
medical articles indexed by PubMed. We  keep only those MeSH terms that are found within the PICO spans. The procedure is applied to all titles and abstracts in our collection.

\paragraph{\textbf{TREC-COVID test set}}

TREC-COVID is a TREC retrieval task addressing literature retrieval related to the  COVID-19 pandemic \cite{roberts-etal-2020-trec,voorhees2020trec}. The topics were developed based on searches submitted to the National Library of Medicine and suggestions from researchers on Twitter. They are representative of the high-level information needs related to the pandemic. Each topic consists of three fields of increasing granularity: a keyword-based query, a natural language question, and a longer descriptive narrative. We use the intermediate-level natural language questions as our queries. The first round of TREC-COVID introduced a set of 30 topics, and was based on the 10 April 2020 release of CORD-19. The most recent, fifth, round includes an additional set of 20 topics, and exploits the 16 July 2020 release of CORD-19. 

The relevance judgments in TREC-COVID were assigned by individuals with biomedical expertise, based on a pooling approach in which only the top-ranked results from different submissions are assessed. A document is judged as `relevant', `partially relevant', or `not relevant'. 
We consider the first two labels as \textit{relevant} in our experiments.


\subsection{Search and visualisation}\label{sec:search}
In COVID-SEE, the CORD dataset is stored as a graph database (neo4j). Queries are executed 
against the full-text index over the titles and abstracts of the papers. We employ a very simple BM25 baseline which while not state of the art represents a standard (strong) baseline for IR tasks \cite{kamphuis2020bm25,trotman2014improvements}. The search capability in neo4j is powered by the Apache Lucene \cite{lucene}, 
with a default VSM scoring function based on tf-idf. We limit the output of the IR search to 1000 hits. 


The relational concept selection organises the identified MeSH concepts into PICO categories, and shows how they interact in a Sankey diagram (Figure~\ref{fig:sankey}). The diagram displays relations between pairs of concepts, where the relations `carry' the corresponding documents, and the  strength of a relation corresponds to the number of documents in which that concept pair is attested. The user interface for the relational concept selection supports varying levels of conceptual match. All our experiments involve MeSH terms of granularity 1, meaning that terms belonging to varying levels of the MeSH hierarchy get mapped to a common level, which is the one of the 16 topmost MeSH categories in combination with a tree descriptor. For example, \textit{Canada} (MeSH tree number: Z01.107.567.176) is mapped (truncated) to \textit{Geographic Locations} (Z01), and \textit{Hydroxychloroquine} (D03.633.100.810.050.180.350) to \textit{Heterocyclic Compounds} (D03).

\subsection{Evaluation details}\label{sec:eval}
We evaluate the retrieval runs with the standard TREC script \cite{treceval}. We report \textit{precision}, 
based on the retrieved documents $R$ that were either judged as relevant (\textit{rel}) or irrelevant (\textit{irrel}), or were not judged at all in the TREC test set (\textit{unj}):
\begin{math}
\frac{|R^{\text{rel}}|}{|R^{\text{rel}}| + |R^{\text{irrel}}| + |R^{\text{unj}}|}
\end{math}.
We additionally report a score that ignores the unjudged documents, referred to as \textit{precision$^{judg}$}.

Note that there is no ranking in the relational concept selection, 
since documents are shown without ordering or scoring. Hence, we do not use evaluation metrics that are rank-dependent.

\section{Analysis}
\subsection{Relevance of documents in the relational concept selection}\label{sec:rel}
The relational concept selection implicitly acts as a filter for the documents retrieved by the IR engine, as some document abstracts do not contain any identifiable P--I or I--O pairs. Overall, the median percentage of documents (across all queries) containing PICO-typed concept relations 
after pre-processing is 13\% (with the minimum and maximum for a given query being 8\% and 28\%, respectively).
To understand what documents are selected and the retrieval quality  of that selection, we carry out several experiments.

\begin{figure}[t!]
\centering
\centering
\includegraphics[width=0.95\linewidth]{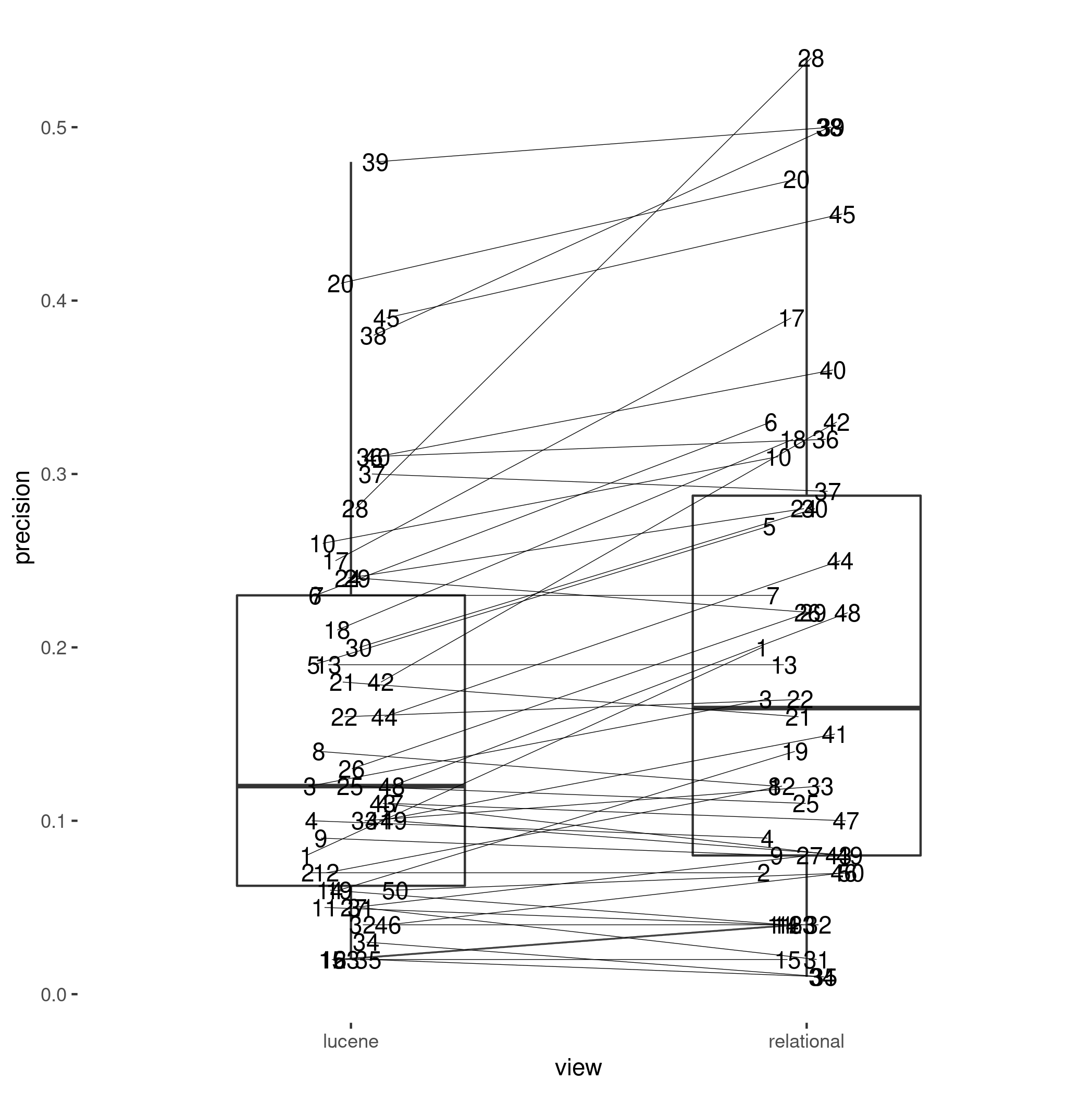}
\caption{A box plot comparing the precision (section \ref{sec:eval}) between the result sets retrieved for each query by the search engine (lucene) and the relational concept selection. The numbers represent the TREC-COVID query identifiers.}
\label{fig:acc_lucene_relational}
\end{figure}

We first compare the precision scores of documents in the raw retrieval results vs.\ documents in the filtered collection, across all 50 queries, as detailed in Figure~\ref{fig:acc_lucene_relational}. The median score for documents in the relational concept selection is higher (0.17) than for the raw search results (0.12). The three queries with the highest precision are as follows, where the last query (28) achieves the highest improvement over raw search results (0.28 $\rightarrow$ 0.54):
\begin{itemize}
    \item[(38)] \textit{What is the mechanism of inflammatory response and pathogenesis of COVID-19 cases?}
    \item[(39)] \textit{What is the mechanism of cytokine storm syndrome on the COVID-19?} 
    \item[(28)] \textit{What evidence is there for the value of hydroxychloroquine in treating Covid-19?}
\end{itemize}
These queries can all be seen as precise statements about the required Problem/Population (\textit{inflammatory response}, \textit{cytokine storm syndrome}, \textit{Covid-19}, respectively) and Intervention (\textit{hydroxychloroquine} in query 28; the other two queries lack this criterion).

\subsection{Role of unjudged documents}\label{sec:unjudged}
The calculation of precision (Section \ref{sec:eval}) includes in the denominator the unjudged documents $R^{unj}$. Since we have little \textit{a priori} intuition about the nature and the relevance of these documents, or whether presenting them to the user can be favourable or not, we manually analysed a sample of them, randomly selecting ten queries and picking for each the first 50 unjudged documents.\footnote{Based on our document ranker.} 
We then checked the abstract of each document to decide whether it is relevant to the query. The results are shown in Table \ref{tab:unjudged}. Although some unjudged documents are relevant  (cf.\ \cite{zobel_1998}), for all but one query the precision of the unjudged documents is far below the prior probability based on TREC-COVID judgements. We therefore infer that including fewer unjudged documents in the results should be favourable. This is in line with the view of the unjudged documents found in the literature \cite{sanderson2010test}.

We observe that the increased precision of results in the relational concept selection is largely due to including fewer unjudged documents: the proportion of retrieved unjudged documents is high in the raw search results ($\mu$=0.74, min=0.5, max=0.94, n=50), but somewhat lower in the relational selection ($\mu$=0.65, min=0.4, max=0.9, n=50). In practice, this means that the user can expect to encounter---per 100 documents retrieved---9 unjudged documents less in the relational view compared to the IR list.

While the number of hits is on average close to 1000 for the search engine, the relational concept view typically captures only around 15\% of the retrieved articles ($\mu$=148 documents, min=78, max=278, n=50). The PICO view filters from the IR search list without any particular order, and in a way that is not guided by the original search ranking.


\begin{table}[t]
\centering
\begin{tabular}{SSSSS}
\toprule
{query} & {\# rel.} & {P} & {P$^{judg}$} & $\Delta$\\
\midrule
2  & 1 & 0.02 & 0.27 & -0.25\\
11 & 3 & 0.06 & 0.17 & -0.11\\
15 & 1 & 0.02 & 0.09 & -0.07\\
18 & 10 & 0.20 & 0.89 & -0.69\\
19 & 4 & 0.08 & 0.45 & -0.37\\
22 & 13 & 0.26 & 0.56 & -0.3\\
26 & 8 & 0.16 & 0.67 & -0.51\\
34 & 8 & 0.16 & 0.06 & 0.10\\
38 & 20 & 0.40 & 0.85 & -0.45\\
47 & 3 & 0.06 & 0.36 & -0.30\\
\bottomrule
\end{tabular}
\caption{Summary of manual annotations of relevance on top-50 unjudged documents from TREC-COVID for each query. \textit{P} refers to our estimated precision among these unjudged documents in TREC-COVID; \textit{P$^{judg}$} is the precision obtained on judged documents only (whether relevant or irrelevant, but excluding the unjudged ones); $\Delta$ quantifies the difference in the reported precision scores.}
\label{tab:unjudged}
\end{table}

\subsection{Exploring the relational concept selection}
\paragraph{\textbf{The nature of queries and their fit to the relational selection}}\label{sec:fittoquery}
Given the design of COVID-SEE in which \textit{all} P- and I-typed concepts (as well as I- and O-typed concepts) are considered to be related if they co-occur within the same abstract without any further constraints, there is no guarantee that the P--I and I--O pairs truly correspond to the initial query. The query may not clearly state any PICO criteria, or the abstract may introduce PICO elements that are not mentioned in the query. Therefore, we would like to better understand what proportion of PICO-typed MeSH terms (or \textit{nodes} of the graph) in the relational concept selection match the PICO criteria expressed in a query. Since the TREC-COVID queries are not annotated with PICO criteria or MeSH terms, we manually annotated all 30 queries from round 1 of the shared task with this information.

The results in Table \ref{tab:queryrel} show that the percentage of nodes that match the query directly is generally low, on average around 7\% for Population and Intervention, and 12\% for Outcome. We find a moderate correlation between the proportions of Intervention concepts fitting the query and the precision scores ($\rho$=0.53), but the number of queries with a clearly stated Intervention is small ($n$=12).

\begin{table}[t!]
\centering
\begin{tabular}{l Sccc}
\toprule
type & $\mu$ & {min} & {max} & $n$ \\
\midrule
P & 7 & 1 & 14 & 20\\
I & 7 & 3 & 17 & 12 \\
O & 12& 3 & 18 & 4 \\
\bottomrule
\end{tabular}
\caption{Percentage of PICO-typed MeSH concepts included in the relation concept selection that are directly relevant to the query. $\mu$ is the mean over $n$ queries (not all queries express PICO criteria).}
\label{tab:queryrel}
\end{table}

\paragraph{\textbf{Extent of grouping}} 
We next explore to what extent the documents represented in the relational concept selection are grouped (pooled) together under the same relation, and whether more evidence of grouping suggests a higher precision or a better fit to the query. 
Analysis (not shown) 
reveals that: (a) most relations are sparse, only corresponding to a single document; (b) the ratio of relations representing more than one document varies between 0.17 and 0.47, depending on the query;
and (c) for some queries, relations cover a large number of documents, e.g.\ for query 7, some relations contain more than 25 documents, whereas for query 8, fewer relations cover a large number of documents. The correlation between the number of documents in a relation and the precision scores for each relation (across all queries) reveals a slight positive correlation, with $\rho$=0.22 on 85,287 relations. 

\section{Further discussion \& conclusion}

From 
this analysis, we can conclude that a query processing approach that directly relies on matching a PICO-based structured query, such as the semantic search available as an alternative search strategy in COVID-SEE as well as in SciSight \cite{Hope2020SciSightCF} and DOC Search \cite{docsearch}, is unlikely to be effective for the TREC-COVID queries.
This reflects the fact that these queries  
are generally quite open-ended \cite{voorhees2020trec}, and in many cases do not clearly follow the PICO structure.
The finding is also consistent with previous research examining the impact of explicitly using defined PICO criteria  in search strategies for systematic reviews, which found that the inclusion of some PICO elements, specifically Outcome, significantly reduced recall~\cite{frandsen2020}.


However, our analysis of information presented in the selection of documents based on PICO categories and medical concepts suggests some benefits of integrating PICO. 
Specifically, the user is more likely to be exposed to more relevant documents, and fewer unjudged documents, compared to the raw search results. Additionally, the user is more likely to find the PICO concepts related to the initial query in those relations that cover a larger number of articles. These stronger relations then tend to represent to a greater extent the articles relevant to the query. 
A limitation is that different queries express the PICO concepts in various degrees, 
which affects the selection of articles and their overall relevance.


\newpage
\bibliographystyle{ieeetr}
\bibliography{anthology,emnlp2020}

\end{document}